# A Comprehensive Evaluation Study on Risk Level Classification of Melanoma by Computer Vision on ISIC 2016 – 2020 Datasets

Chengdong Yao
School of Computer Science
University of Technology Sydney
Australia
chengdong.yao-1@student.uts.edu.au

Mukesh Prasad
School of Computer Science
University of Technology Sydney
Australia
mukesh.prasad@uts.edu.au

Kunal Chaturvedi
School of Computer Science
University of Technology Sydney
Australia
kunal.chaturvedi@uts.edu.au

*Abstract* — Skin cancer is the most common type of cancer. Specifically, melanoma is the cause of 75% of skin cancer deaths, although it is the least common skin cancer. Better detection of melanoma could have a positive impact on millions of people. The ISIC archive contains the largest publicly available collection of dermatoscopic images of skin lesions. In this research, we investigate the efficacy of applying advanced deep learning techniques in computer vision to identify melanoma in images of skin lesions. Through reviewing previous methods, including pre-trained models, deep-learning classifiers, transfer learning, etc., we demonstrate the applicability of the popular deep learning methods on critical clinical problems such as identifying melanoma. Finally, we proposed a processing flow with a validation AUC greater than 94% and a sensitivity greater than 90% on ISIC 2016 – 2020 datasets.

## I. Introduction

In recent years, melanoma has become more common. This is thought to be the result of exposure to the strong sun while on holiday. About 16,000 new cases of melanoma are diagnosed in the UK each year, and more than 2,300 people die from melanoma [1]. As with other cancers, early and accurate detection, perhaps with the help of data science, can make treatment more effective. Melanoma is fatal, but if detected early, most melanoma can be cured with minor surgery. Previous work has summarized the visual features that can be used to artificially distinguish between normal mole and melanoma [2]. Image analysis tools for automatic melanoma diagnosis will improve dermatologists' diagnostic accuracy.

Although deep learning has achieved new performance highs, they require large, high-quality annotated datasets. However, like other medical image datasets, ISIC has only a low number of positive samples and only image-level noisy annotations for training data. In the presence of these dataset defects, even the most advanced models may not be able to extend real-world clinical environments. To handle this challenge, we need to actively seek solutions and adopt diversified and effective techniques.

Our findings and contributions include:

1) A detailed exploration of the ISIC 2016-2020 datasets is done, including the distribution of positive and negative samples, and investigated how peers alleviate the problem of sample imbalance.

2) Many experiments have been carried out to verify the performance of different backbones. Referring to some previous work, many different model architectures pre-trained on Image Net are applied to our solution with experiments.

3) Summarized the improvements that all the techniques discussed in this report can bring to the current task and proposed a processing flow that the validation AUC is over 94% and the sensitivity is over 90%.

## II. Literature Review

We focus on machine learning methods that are used to detect melanoma. A. Naeem et al. selected 55 of the 5,112 research in six databases, including IEEE and Medline [3]. Their main objective is to collect state-of-the-art research, identify the latest trends, challenges and opportunities in melanoma diagnosis, and explore existing solutions for melanoma testing using deep learning diagnosis. In addition, they summarized various existing melanoma detection solutions.

This study focuses on the proposed methods that use ISIC 2016-2020 datasets. We start the review by finding the related papers. We looked at several studies that include finetuning using pre-trained models, fully convolution networks, ensemble learning, and adding hand-crafted features. However, we cannot involve all the related studies because some of the research is either conducted on undisclosed internal datasets or does not provide enough detailed information to reproduce experiments. Based on comparable requirements, here are some of the research that we selected for our investigation.

TABLE I. THE LITERATURE WE INVESTIGATED AND THE TECHNIQUES THEY USED

| Publication Title | Techniques Used |
| --- | --- |
| Skin lesion classification using hybrid deep neural networks [4] | AlexNet; VGG16; ResNet-18 |
| Combining deep learning and hand-crafted features for skin lesion classification [5] | Deep Neural Network |
| Classification of dermoscopy patterns using deep convolutional neural networks [6] | Gradient Descent Algorithm |
| Automated melanoma recognition in dermoscopy images via very deep residual networks [7] | Deep Residual Network |
| Towards automated melanoma detection with deep learning: Data purification and augmentation [8] | Transfer Learning |
| Automated skin lesion classification using ensemble of deep neural networks in ISIC 2018: Skin lesion analysis towards melanoma detection challenge [9] | Ensemble Deep Neural Network |
| Fusing fine-tuned deep features for skin lesion classification [10] | Ensemble Deep Neural Network |
| An image-based segmentation recommender using crowdsourcing and transfer learning for skin lesion extraction [11] | VGG16; ResNet-50; Crowd Sourcing |
| Hybrid fully convolutional networks based skin lesion segmentation and melanoma detection using deep feature [12] | Fully Convolutional Network |
| Use of neural network-based deep learning techniques for the diagnostics of skin diseases [13] | Transfer Learning |
| Convolutional neural network algorithm with parameterized activation function for melanoma classification [14] | Parameterized Activation Function |
| Multi-model deep neural network based features extraction and optimal selection approach for skin lesion classification [15] | Transfer Learning |
| Melanoma detection using adversarial training and deep transfer learning [16] | Transfer Learning |
| Soft Attention Improves Skin Cancer Classification Performance [17] | Soft Attention |

A lot of extensive research has been carried out in this field, and some new techniques and algorithms have been relying on deep learning to diagnose melanoma. This can be divided into four main categories:

1) Using pre-trained convolution neural network to extract high-level numerical features.

   A. Mahbod et al. use pre-trained CNN to study the classification of skin lesions and implement pre-trained AlexNet and VGG-16 architecture to classify skin lesions in their algorithm to extract different features from dermatoscopic images [4]. A. Soudani et al. use the hybrid technology of hybrid pre-training architecture (VGG16 and ResNet50) to extract features from the convolution part [11]. A classifier with an output layer is designed composed of five nodes. These nodes represent the categories of segmentation methods and predict the most effective skin lesion detection and segmentation techniques in any image data.

2) Using a combination of hand-crafted features and network features.

   T. Majtner et al. introduced a melanoma recognition system that includes manual features and comprehensive features [5]. In their study, deep learning and support vector machine were used for melanoma diagnosis. For grayscale images, support vector machines are used to extract features, and for the convolution of original colour images, neural networks are used to generate likelihood scores. The result is based on the high score. S. Demyanov et al. proposed a method of using CNN to perceive the two-mode forms (typical network and regular ball) in the dermatoscopic image [6].

3) Using residual blocks instead of traditional convolution layers as mentioned in the fully convoluted network.

   L. Yu et al. proposed a method for diagnosing melanoma using dermatoscopic images using a depth residual network [7]. The proposed method uses two FCNs that replace the traditional convolution layer with residual blocks, as mentioned in the FCNs architecture. In addition, the proposed method generates a grading map from a dermatoscopic image to a segmented skin lesion. Regions of interest containing skin lesions have been resized, cropped, and metastasized for melanoma classification. In addition, K. Jayapriya et al. adopted a hybrid framework containing two FCNs (VGG 16 and GoogleNet) [12]. Deep residual networks and hand-made tools are used for classification.

4) Using ensemble learning within the same architecture or between different architectures.

   Many studies use ensemble deep learning techniques for melanoma classification, such as the collaborative deep learning model used by A. Milton et al. [9], which was tested on the benchmark data set of ISIC 2018. Identify skin lesions from dermatoscopic images. However, A. Mahbod et al. proposed a technology based on deep collaborative learning [10], which is designed through the integration of intra-and inter-architecture networks of a unified convolution neural network (CNN). This is an entirely automatic and instinctive computerization process.

III. DATA PREPROCESSING

We make a detailed exploration of the ISIC data sets used, including the distribution of positive and negative samples, and investigate how peers alleviate the problem of class imbalance.

A. Data Distribution

Because several studies use different datasets, we explore all the datasets from ISIC 2016 to ISIC 2020 in depth. It is worth noting that although they have been released independently, ISIC 2017 is included in ISIC 2018 and ISIC 2018 is included in ISIC 2019. ISIC 2016 and ISIC 2020 are independent datasets. So, it is only needed to combine ISIC 2020 with ISIC 2019 and ISIC 2016 to get all the data. This can be verified by the data in Table II below.

As shown in Table II below, ISIC 2020 is a highly imbalanced dataset with a positive rate of less than 2%. Even if we use all the data, less than 9% of the samples are positive, which will significantly impact the model's recall rate. The most serious difficulty in classifying such datasets is the long tail effect caused by data imbalance, which can be alleviated by transfer learning or ensemble learning. There are also the ways many works use, such as [11, 16].

TABLE II. DATASET ISIC 2016 – ISIC 2020

| Dataset | Positive Samples | | Negative Samples | |
|---|---|---|---|---|
| | Count | Ratio | Count | Ratio |
| ISIC 2016 [18] | 173 | 19.2% | 727 | 80.8% |
| ISIC 2017 [19] | 374 | 18.7% | 1,626 | 81.3% |
| ISIC 2018 (Incl. 2017) [20] [21] | 1,113 | 11.1% | 8,902 | 88.9% |
| ISIC 2019 (Incl. 2018) [19] [21] [22] | 4,522 | 17.9% | 20,809 | 82.1% |
| ISIC 2020 [23] | 584 | 1.8% | 32,542 | 98.2% |
| ALL | 5,279 | 8.9% | 54,078 | 91.1% |

*B. TFRecord samples*

The original data are images in JPEG format with different resolutions, and the labels are provided in the form of CSV tables, all of which are publicly accessible. Ref. [24] on the Kaggle platform have sorted it into TFRecord format and provides different resolutions to choose from, which is more convenient for our research. The following shows some positive (Figure 2) and negative (Figure 1) samples randomly selected from TFRecord.

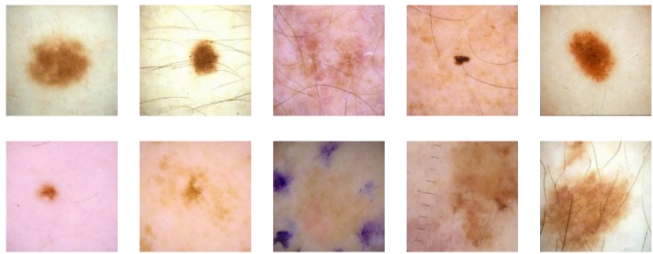

Fig. 1. Some negative (normal mole) samples

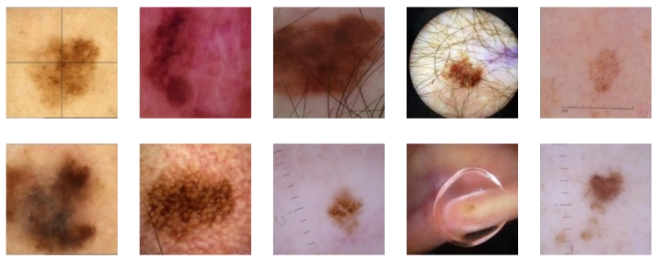

Fig. 2. Some positive (melanoma) samples

## IV. METHODOLOGY

We use many different methods to improve the performance of learning algorithms to detect melanoma. In addition to the general augmentation methods commonly used by predecessors, we also explore the efficacy of Coarse Drops in our data and tasks. Up-sampling of positive samples has been proved to effectively improve the performance of the model and alleviate the problem of sample imbalance. This chapter will introduce our entire processing flow and related methodologies.

*A. Processing Flow*

Since we are using data based on TFRecord format, we can divide it into K-fold directly after loading these data. The ISIC dataset contains images as well as metadata. The images are first augmented through a series of operations and then fed to the CNN-based backbone. Metadata cannot be directly input into CNN, so it needs to be processed differently. The encoded metadata will be fed into the ensemble model, and the feature vectors extracted by the CNN backbone will also be input for training. This process is repeated K times so the model can use all the data. The output of the ensemble model is the final output, which will be collected and evaluated.

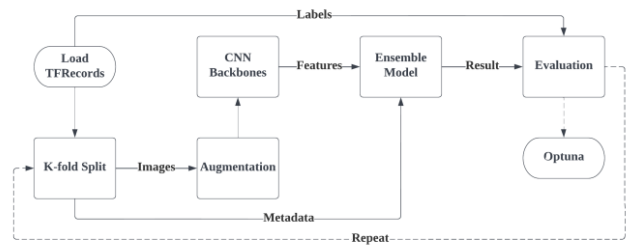

Fig. 3. The whole processing flows

The flowchart (Figure 3) above demonstrates the whole processing flow. We will repeat the above process with different hyperparameters and backbones to compare performance and efficiency under different configurations. In all the experiments, due to the need for the control variable, the hyperparameters take the same value as far as possible, which is probably not optimal. Therefore, hyperparameter optimization (Optuna) will be implemented to further improve the overall performance.

*B. Augmentation*

Deep neural networks need a lot of training data to achieve good performance. To use a small amount of training data to build a powerful and robust image classifier, image augmentation technology is usually needed to improve the generalization ability of the network. Image augmentation technology artificially creates training images through the combination of many different processing methods, such as random flipping, clipping, shifting and rotation [26].

The following flowchart (Figure 4) shows our process of image augmentation.

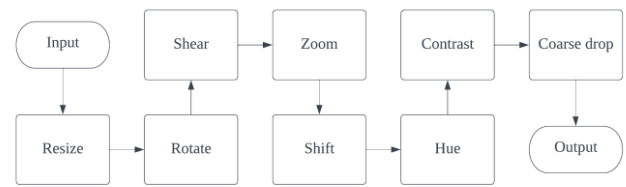

Fig. 4. Image augmentation steps

The input image size is a hyperparameter, such as 128, 192, 256, etc. The first step in image augmentation is to resize it to the desired value. Because the random rotation of the image does not affect whether it is melanoma, this is what we will do next. Similarly, shearing, zooming, shifting and other operations will be successively applied to the image and then adjust the hue and contrast. These operations will affect the content of the original image, so each operation will have a corresponding hyperparameter to control its degree. These are commonly used image augmentation methods, mainly to imitate the data differences under different imaging conditions.

Coarse dropout augmentation is a technique that prevents overfitting and improves generalization [27]. It randomly removes some rectangles from the training image. This is different to the dropout layer in traditional CNN. Deleting a part of the image leads to challenging the model to focus on the whole image because the model never knows which part of the image will disappear. The following Figure 5 shows some image samples processed by Coarse dropout.

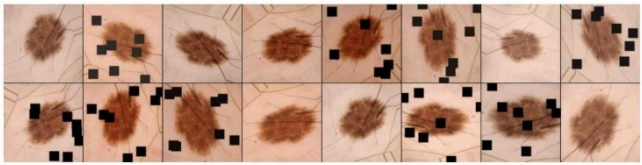

Fig. 5. Some augmented image samples

### C. Upsampling

Since all the positive samples of ISIC 2016-2020 add up to only 5,279, less than 9%. This is not enough for deep learning, even if image augmentation is used. We also obtained an additional 3976 positive sample images of high quality by up-sampling in the augmented image set. Adding these upsampled images may help to alleviate the problems caused by data imbalance. As shown in table III below, the proportion of positive samples has now risen to nearly 15%, the imbalance has been alleviated.

TABLE III. DATASET WITH UP SAMPLED IMAGES

| Dataset | Positive Samples | | Negative Samples | | Total Samples |
|---|---|---|---|---|---|
| | Count | Ratio | Count | Ratio | |
| ISIC 2016-2020 | 5,279 | 8.9% | 54,078 | 91.1% | 59,357 |
| With up sampled | 9,255 | 14.6% | 54,078 | 85.4% | 63,333 |

### D. Backbone

In computer vision, many deep learning models are provided with pre-training weights, which are called backbone. We also use different backbones for feature extraction to study the kind of backbone suitable for this task. All these backbones are pre-trained on ImageNet and can be used directly in TensorFlow. The basic information about these models is shown in the following table.

TABLE IV. BASIC INFORMATION ON SELECTED BACKBONES [36]

| Backbone | Size (MB) | Parameters (M) | Depth (layer) | GPU infer speed (fps) |
|---|---|---|---|---|
| Xception [28] | 88 | 22 | 81 | 124 |
| VGG 16 [29] | 528 | 138 | 16 | 239 |
| ResNet 101 V2 [30] | 171 | 44 | 205 | 186 |
| Inception V3 [31] | 92 | 23 | 189 | 145 |
| IncepResNet V2 [32] | 215 | 55 | 449 | 100 |
| MobileNet V2 [33] | 14 | 3 | 105 | 264 |
| DenseNet 169 [34] | 57 | 14 | 338 | 159 |
| EfficientNet B6 [35] | 166 | 43 | 360 | 25 |

We also refer to some previous work, apply many different model architectures pre-trained on Image Net to our solution, and compare their performance and efficiency with experiments. Ensemble learning and hyperparameter optimization have an important impact on the performance of the solution. In particular, ensemble learning with the GBDT strategy directly improves the performance of the model by 4%.

### E. Evaluation

To compare directly with the work of others, we use the AUCROC (Area Under the Curve of the Receiver's Operating Characteristic) of positive samples recommended by the dataset as our performance evaluation function. In addition, we also used the sensitivity (recall rate) of positive samples to compare with the state-of-art. We mainly use stratified K-fold cross-validation to divide the data. Stratified sampling is required because this is an imbalanced dataset. Partitioning that is applied directly may significantly change the data distribution and affect the accuracy of the testing. Because the K-fold cross-validation is very slow, we use a more straightforward partition method to speed up the experiment when we try different backbones on a large scale. It has been proved (see the next chapter for details) that the result of the simple partition method of dividing the data into training, validation and testing sets is almost the same as that of the corresponding K-fold cross-validation.

### F. Ensemble and Hyperparameter

The ensemble learning method uses various machine learning algorithms to achieve better prediction performance than any single algorithm [37]. It consists of a specific set of limited alternative models that usually allow for more flexible architectures in these alternatives. There are many ways to achieve this. We will first use the different backbones to extract image features and study the performance of Blending, MLP (Multi-Layer Perceptron) and GBDT (Gradient Boosting Decision Tree) in our task. Details will be described in the specific experiments in the next chapter.

It is crucial to adjust hyperparameters in deep learning. This process can train models with better performance more scientifically. Generally, the training state of the model is judged by observing the value of the loss function or the performance of the validation set, and different hyperparameters are tried to improve the model's performance. In the past, grid search and random search are commonly used hyperparameter optimization methods. Currently, Optuna [38] is a well-known hyperparameter optimization tool, that uses advanced algorithms to sample hyperparameters and effectively prune hopeless experiments to achieve very efficient hyperparameter optimization.

## V. EXPERIMENTS

We conducted a total of 29 different experiments, divided into 15 groups. Several configurations have been tested, and our model has been gradually improved. The following are the detailed results of the methods. We will describe it in four parts, single models, ensemble models, transfer learning and hyperparameter optimization.

## A. Clustering analysis

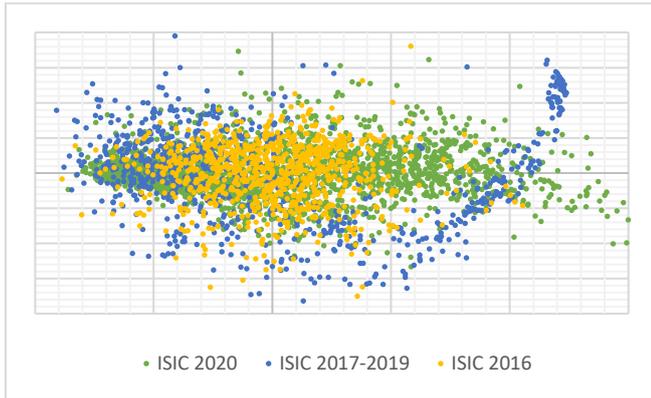

Fig. 6. Visualisation of dimension reduction results

Since our work is based on datasets from different periods, we have to prove that these datasets can be used together without disagreement. After applying TSVD (Truncated Singular Value Decomposition) [25] to the image, the result of dimension reduction of these datasets is shown in Figure 6 above. It can be seen that the data from these sources are concentrated, and there is no simple way to separate them. This means that their distribution is uniform and unambiguous. Therefore, it makes sense to combine these datasets together.

## B. Single Models

TABLE V. EXPERIMENT 1-15 (DIFFERENT SINGLE MODELS)

| Experiment | Model | Dataset | Input size | Train AUC | Valid AUC |
|---|---|---|---|---|---|
| #1 | Xception | ISIC 2020 | 128 | 95.25% | 83.81% |
| #2 | Xception | ISIC ALL | 128 | 96.06% | 84.10% |
| #3 | VGG 16 | ISIC ALL | 128 | 94.71% | 85.43% |
| #4 | ResNet 101 V2 | ISIC ALL | 128 | 94.93% | 83.38% |
| #5 | Inception V3 | ISIC ALL | 128 | 94.42% | 84.18% |
| #6 | IncepResNet V2 | ISIC ALL | 128 | 96.32% | 83.07% |
| #7 | MobileNet V2 | ISIC ALL | 128 | 89.76% | 79.66% |
| #8 | DenseNet 169 | ISIC ALL | 128 | 95.88% | 84.93% |
| #9 | EfficientNet B6 | ISIC ALL | 128 | 94.64% | 85.84% |
| #10 | VGG 16 | ISIC Up sampled | 128 | 96.40% | 85.53% |
| #11 | EfficientNet B6 | ISIC Up sampled | 128 | 96.41% | 86.34% |
| #12 | VGG 16 | ISIC Up sampled | 192 | 96.05% | 86.03% |
| #13 | EfficientNet B6 | ISIC Up sampled | 192 | 98.29% | 87.37% |
| #14 | VGG 16 | ISIC Up sampled | 256 | 95.82% | 85.27% |
| #15 | EfficientNet B6 | ISIC Up sampled | 256 | 98.86% | 87.26% |

TABLE VI. EXPERIMENT GROUP 5 (CROSS-VALIDATION)

| Experiment | Model | Dataset | Input size | CV AUC | Valid AUC |
|---|---|---|---|---|---|
| #12 | VGG 16 | ISIC Up sampled | 192 | N/A | 86.03% |
| #16 | VGG 16 | ISIC Up sampled | 192 | 86.03% | N/A |
| #13 | EfficientNet B6 | ISIC Up sampled | 192 | N/A | 87.37% |
| #17 | EfficientNet B6 | ISIC Up sampled | 192 | 87.43% | N/A |

The first group of experiments (#1 and #2) as described in Table V aims to prove that the performance of using all datasets of ISIC is better than using ISIC 2020 alone. The experimental results show that the performance of using all data is better in both training and validation. Therefore, all data are used in the following experiments unless otherwise stated.

In the previous chapter, we investigated many backbones commonly used in computer vision. The second group of experiments (#2 - #9) measured these backbones to determine the kind of backbone most suitable for this task. As can be seen from Table V above, the validation AUC of most models falls at about 84%, with the highest being EfficientNet and VGG, which is more than 85%. The experiments of the later order will be carried out on this basis.

Although ISIC ALL contains all available samples, the proportion of positive samples is too low. We retrain the relevant backbone in the up sampled dataset to prove the effectiveness of the up-sampling method. The experimental results (#3 and #9 - #11) show that up-sampling of positive samples can improve the performance of the model, in both EfficientNet and VGG. So, we have enabled the up-sampling of positive samples in the following experiments.

Considering that the input size of 128 may be too small, it is difficult for the model to extract effective features from it. The fourth group of experiments (#10 - #15) studied the effects of different input sizes on the performance of the model, and the input sizes were 128,192 and 256, respectively. As can be seen from the experimental results in Table V above, the input size of 192 is the most appropriate for this task. The following experiments are all based on the input size of 192.

The above experiments all use simple training, validation and testing partitions to shorten the time of the experiment. The fifth group of experiments (#12 - #13 and #16 - #17) compare the two partition methods to prove that it is feasible. It is proved that when the amount of training data is the same, the results of simple partition and k fold cross validation are similar in the error range. However, we will still use 5-fold cross-validation by default later.

## C. Ensemble Models

TABLE VII. EXPERIMENT 16-20 (DIFFERENT ENSEMBLE MODELS)

| Experiment | Model | Dataset | Input size | Train AUC | CV AUC |
|---|---|---|---|---|---|
| #16 | VGG 16 | ISIC Up sampled | 192 | 96.13% | 86.03% |
| #17 | EfficientNet B6 | ISIC Up sampled | 192 | 98.29% | 87.43% |
| #18 | Blending Ensemble | ISIC Up sampled | 192 | 97.68% | 87.30% |
| #19 | MLP Ensemble | ISIC Up sampled | 192 | 98.36% | 89.10% |
| #20 | GBDT Ensemble | ISIC Up sampled | 192 | 98.81% | 91.34% |

TABLE VIII. EXPERIMENT GROUP 10 (COARSE DROP AUGMENT)

| Experiment | Model | Dataset | Input size | Train AUC | CV AUC |
|---|---|---|---|---|---|
| #19 | MLP Ensemble | ISIC Up sampled | 192 | 98.36% | 89.10% |
| #21 | MLP Ensemble | ISIC Coarse drop | 192 | 95.06% | 88.61% |

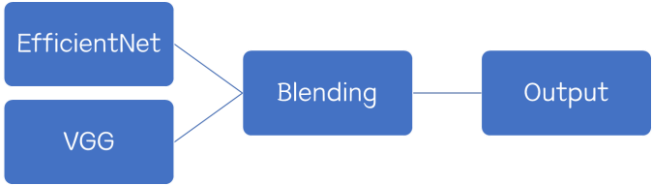

Fig. 7. Schematic diagram of Blending Ensemble

Blending is a simple ensemble learning strategy which directly weighs and averages the output of different models. We blend and compare the output of EfficientNet and VGG (Figure 7). The experimental results (#16 - #18) show that the blending model's performance decreases by 0.1%, which indicates that blending does not improve this task. This suggests that we need to use more effective ensemble learning strategies.

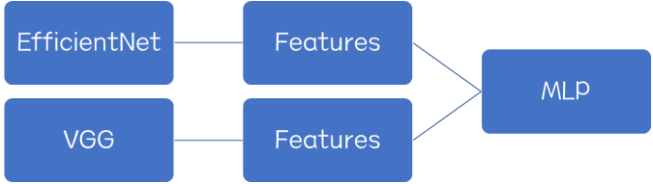

Fig. 8. Schematic diagram of MLP Ensemble

Integrating two trained models through a multi-layer perceptron (MLP) is a better solution. In the seventh group of experiments (#16 - #17 and #19), the features in Figure 8 refer to the input of the last fully connected (linear/dense) layer. This time, the performance of the ensemble model has been improved in both training and validation. This shows that MLP can take advantage of the diversity of EfficientNet and VGG to improve model performance.

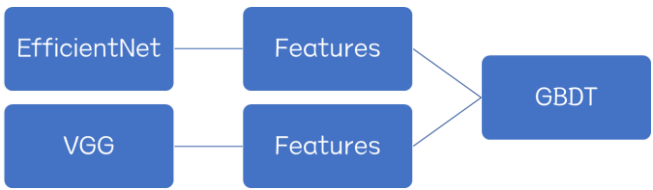

Fig. 9. Schematic diagram of GBDT Ensemble

Gradient boosting decision tree (GBDT) is often better for using diversity and does not require tedious hyperparameter tuning. So, in the eighth group of experiments (#16 - #17 and #20), we kept the other configurations like before and replaced MLP with GBDT (Figure 9). From the experimental results, we can see that using GBDT for ensemble learning can improve the performance of the model by up to 4%. This was slightly higher than expected and enabled our validation AUC to exceed 90%.

To more intuitively compare the effects of different ensemble learning strategies on the model's performance, the ninth group of comparisons (#18 - #20) is created. As shown in Table VII, GBDT is 2% better than MLP and 4% better than blending. Ensemble learning is beneficial to this task, which brings more benefits than adding image augmentation. However, GBDT cannot be used in the end-to-end neural network, so it is inconvenient to use.

In image classification, the coarse drop is not a commonly used image augmentation technique, which may have some disadvantages. The 10th group of experiments (#19 and #21) investigated the impact of a coarse drop on the model's performance. When the coarse drop is turned on, the model's performance degrades, indicating a more significant adverse impact on the model. So, we ended up shutting down that to improve performance.

D. Transfer Learning

TABLE IX. EXPERIMENT GROUP 11 (IN OUR DATASET)

| Experiment | Model | Dataset | Input size | Train AUC | CV AUC |
|---|---|---|---|---|---|
| #22 | Hasib's [16] | ISIC Up sampled | 192 | 95.95% | 88.33% |
| #23 | Datta's [17] | ISIC Up sampled | 192 | 95.59% | 87.81% |
| #19 | MLP Ensemble (Ours) | ISIC Up sampled | 192 | 98.36% | 89.10% |

TABLE X. EXPERIMENT 20 AND 22-28 (DIFFERENT TRANSFER LEARNING MODELS)

| Experiment | Model | Dataset | Input size | Sens. | CV AUC |
|---|---|---|---|---|---|
| #20 | GBDT Ensemble | ISIC Up sampled 5F | 192 | 82.70% | 91.34% |
| #22 | Hasib's [16] | ISIC 2016 | 192 | 91.8% | 81.2% |
| #23 | Datta's [17] | ISIC 2017 | 192 | 91.6% | 90.4% |
| #24 | GBDT Ensemble | ISIC Up sampled 10F | 192 | 84.48% | 93.57% |
| #25 | GBDT Ensemble | ISIC Up sampled 15F | 192 | 85.98% | 93.70% |
| #26 | MLP Ensemble (Ours) | ISIC 2016 | 192 | 93.3% | 83.5% |
| #27 | MLP Ensemble (Ours) | ISIC 2017 | 192 | 92.6% | 86.7% |
| #28 | GBDT Ensemble | ISIC Up sampled 15F + metadata | 192 | 88.17% | 93.73% |

Although EfficientNet and VGG have been pre-trained on ImageNet, which is not a skin cancer dataset. In the 11th group of experiments (#19 and #22 - #23), we tried to add some relevant pre-trained models to provide some related features. Using MLP as the ensemble, the addition of two related models did not improve but led to performance degradation. We think this is because EfficientNet and VGG are good enough, and adding relevant models leads to additional deviation.

To verify this conjecture, we need to retrain our solution with the same dataset they use. Table X above (#22 - #23 and #26 - #27) compares the performance of our models on their dataset and their own. Our solution is better than Hasib's, with 2% higher validation AUC and sensitivity. Our AUC lower than Datta's may be due to our lack of hyperparameter optimization, because our model is more sensitive. This explains why adding them to the ensemble model does not help.

The previous model's performance validation uses 5-fold cross-validation (K=5), which results in 20% of the data not being trained. The 13th group of experiments (#20 and #24 - #25) further configured the K to 10 and 15 allowing the model to train for a longer time with more data. The experimental results show that the efficiency of K=10 is higher, but the performance of K=15 will still be slightly improved. This means that more data and longer training time can still help.

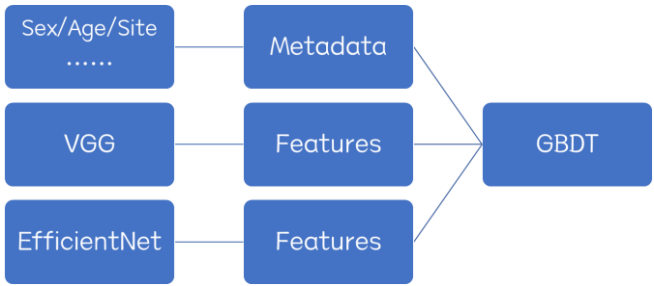

Fig. 10. Schematic diagram of adding metadata

The ISIC dataset contains images and metadata such as gender, age, sites, etc., which may also be helpful. After adding these metadata to the GBDT ensemble model (Figure 10), the results of the 14th group of experiments (#25 and #28) confirmed this conjecture. Although validation AUC has only improved a little, its sensitivity has increased by more than 2%. Therefore, the metadata is of practical help to the task.

*E. Hyperparameter Optimization*

TABLE XI. EXPERIMENT GROUP 15 (DEFAULT VS OPTUNA)

| Experiment | Model | Dataset | Input size | Sens. | CV AUC |
|---|---|---|---|---|---|
| **#28** | GBDT Ensemble (Default) | ISIC Up sampled 15F + metadata | 192 | 88.17% | 93.73% |
| **#29** | GBDT Ensemble (Optuna) | ISIC Up sampled 15F + metadata | 192 | 90.88% | 94.13% |

At the end of all experiments, we use Optuna to optimize the optimal solution. This is a necessary step. To control variables, all the above experiments use the same hyperparameter configuration, which is probably not the optimal one. After optimizing key hyperparameters such as learning rate and depth, the validation AUC has exceeded 94%, and the sensitivity has exceeded 90%. Compared with 83% of AUC at the beginning of the research, we have achieved an 11% improvement.

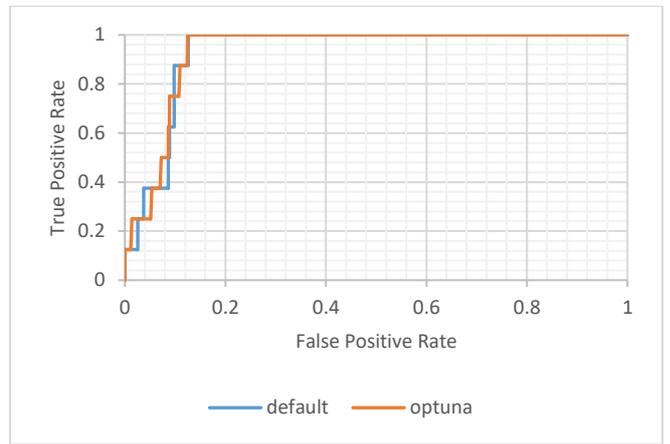

Fig. 11. The ROC curve

## VI. DISCUSSION

Many experiments have been carried out to verify the performance of different backbones. We also compared the transfer learning results using the work of others and the performance of our solution on the dataset they used, which proved that our solution is better. The effects of different ensemble learning strategies and the improvement of model sensitivity caused by hyperparameter optimization are also proved by experiments.

Apart from proving how helpful many techniques commonly used in other tasks can be in this task, the main output is the solution.

It uses EfficientNet B6 and VGG 16 pre-trained on Image Net as the backbone, trained for 15 folds using the up sampled ISIC ALL dataset with an input size of 192. Then the features they extract, the input of the last fully connected (linear/dense) layer, are spliced together with the metadata and are ensembled through a gradient boosting decision tree (GBDT), which is also trained for 15 folds. Figure 12 illustrates all the techniques that have been improved in this task.

From the comparative experiments in the previous chapter, it can be inferred that the incomplete version of our solution exceeds that of Hasib and Datta by about 2% on the same dataset, so it is indeed valuable. Under the appropriate hyperparameter configuration, the validation AUC of our proposed solution can exceed 94%, and the sensitivity can exceed 90%, which is 11% higher than that of the Xception-based baseline.

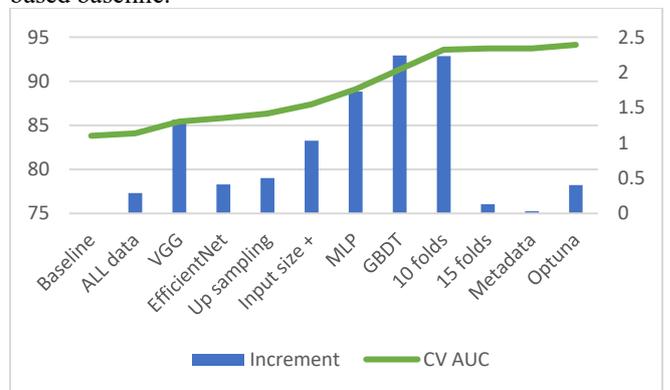

Fig. 12. Increment brought by each technique

## VII. STAKEHOLDER BENEFITS

### A. Doctor

The doctor's time is very precious. Making a diagnosis faster can not only prevent the disease from worsening but also enable doctors to serve more patients. The solution we proposed can be integrated into the computer-aided diagnosis system to help doctors judge the disease more quickly and accurately. It can also be used in the training of doctors to continuously improve their diagnostic ability through timely feedback.

### B. Patient

It is impractical to require ordinary people to tell immediately whether it is a mole or a melanoma. For patients, if they have such an APP on their smartphone, they can directly take photos to identify how likely it is to be melanoma, so they can detect the problem earlier without delaying treatment. If it is just a normal mole, then it is not necessary to go to the hospital. Otherwise, it will occupy medical resources and waste their time.

## VIII. CONCLUSION

We focus on the work that uses ISIC 2016-2020 datasets, starting with a review, finding papers that use this dataset, and skimming through it. Then, a detailed exploration of the ISIC datasets is done, including the distribution of positive and negative samples, and investigated how peers alleviate the problem of sample imbalance.

Many experiments have been carried out to verify the performance of different backbones. We use many different methods to improve the performance of our model. Referring to some previous work, many different model architectures pre-trained on Image Net are applied to our solution and compare their performance and efficiency with experiments.

Finally, we summarized the improvements that all the techniques discussed in this report can bring to the current task and obtained a processing flow that the validation AUC is over 94% and the sensitivity is over 90%. Future research directions are put forward, such as extending to the object detection task and improving the backbone's architecture.

## IX. FUTURE WORK

Considering that we have been able to achieve more than 94% of the validation AUC, the future can go further in two directions, data and architecture.

In terms of data, datasets that are not limited to melanoma can be used in the future, but datasets that contain other common skin cancers. This will not only increase the practicability of the model but also give the model more information because there are other types besides mole and melanoma. The task can not only be a single label classification, if there is mask data, it can also be used as an object detection task. The additional boundary information can make the model more focused on extracting information from specific areas and reduce overfitting.

On the architecture side, although we have experimented with many different backbones and compared their performance and efficiency, none of them is specifically designed for this task. If we can make some small, targeted innovations, instead of using the backbone proposed by others directly, we may be able to achieve better performance or efficiency, or a balance between them.